\documentclass{article}

\usepackage{microtype}
\usepackage{graphicx}
\usepackage{booktabs} 
 \usepackage{amsfonts} 
 \usepackage{amsmath}
 \usepackage{amsthm}
 \usepackage{pifont}
 \usepackage{authblk}
\usepackage{enumitem}
 \usepackage{graphicx}
 \usepackage{hyperref}
 \usepackage{algorithm} 
 \usepackage{algorithmic}

\usepackage{subfig}
\usepackage{amssymb}
\usepackage{setspace}
\usepackage{url}
\usepackage{color}

\newtheorem{thm}{Theorem}

\newtheorem*{defi*}{Definition}

\newenvironment{sproof}{%
 \proof}{\endproof}

\usepackage{hyperref}

\usepackage[accepted]{icml2019}

\icmltitlerunning{Sparse Representation Classification via Screening for Graphs}

\begin{document}

\twocolumn[
\icmltitle{Sparse Representation Classification via Screening for Graphs}


\icmlsetsymbol{equal}{*}

\begin{icmlauthorlist}
\icmlauthor{Cencheng Shen}{equal,to}
\icmlauthor{Li Chen}{equal,goo}
\icmlauthor{Yuexiao Dong}{ed2}
\icmlauthor{Carey Priebe}{ed}

\end{icmlauthorlist}
\icmlaffiliation{to}{Department of Applied Economics and Statistics, University of Delaware}
\icmlaffiliation{goo}{Security and Privacy Research, Intel Labs, Hillsboro 97124}
\icmlaffiliation{ed2}{Department of Statistical Science, Temple University}
\icmlaffiliation{ed}{Department of Applied Mathematics and Statistics, Johns Hopkins University}

\icmlcorrespondingauthor{Cencheng Shen}{shenc@udel.edu}
\icmlcorrespondingauthor{Li Chen}{lichen.jhu1@gmail.com}

\icmlkeywords{Sparse representation classification, stochastic blockmodel, Consistency}

\vskip 0.3in
]



\printAffiliationsAndNotice{\icmlEqualContribution} 

\begin{abstract}
The sparse representation classifier (SRC) is shown to work well for image recognition problems that satisfy a subspace assumption. In this paper we propose a new implementation of SRC via screening, establish its equivalence to the original SRC under regularity conditions, and prove its classification consistency for random graphs drawn from stochastic blockmodels. The results are demonstrated via simulations and real data experiments, where the new algorithm achieves comparable numerical performance but significantly faster. 
\end{abstract}

\section{Introduction}
\label{sec:intro}

Sparse coding is a useful and principled tool in machine learning, due to the theoretical advancement in regularized regression and $\ell 1$ minimization \cite{OsbornePresnellTurlach2000a, OsbornePresnellTurlach2000b, DonohoHuo2001, EfronHastie2004, CandesTao2005, Donoho2006, CandesTao2006, CandesRombergTao2006}, and it is also effective in numerous classification and clustering applications in computer vision and pattern recognition \cite{WrightYangGaneshMa2009, WrightMa2010, YinEtAl2012, YangZhouGanesh2013, ElhamifarVidal2013, ChenShenVogelsteinPriebe2016}.

Here we focus on the sparse representation classification (SRC), which is proposed by \cite{WrightYangGaneshMa2009} to combine sparse encoding with computer vision. This classification scheme exhibits state-of-the-art performance for robust face recognition. It has straightforward implementation and works well for data satisfying the subspace assumption (e.g. face recognition, motion segmentation, and activity recognition). Furthermore SRC is a robust classifier against data contamination and extensive to block-wise SRC and structured data sets \cite{EldarMishali2009,EldarKuppingerBolcskei2010,ElhamifarVidal2012}. In the task of face recognition, where a number of face images $\mathcal{X}=[x_{1},\ldots,x_{n}]$ with the corresponding class labels $\mathcal{Y}=[y_{1},\ldots,y_{n}]$ are considered the training data, the task is to classify a new testing image $x$. SRC identifies a small subset $\widehat{\mathcal{X}}$ in the training data that best represent the testing image, calculates the least square regression coefficients, and computes the regression residual for classification.

The intrinsic mechanism of SRC is not well-understood. A number of literature suggest that neither $\ell 1$ minimization nor the subspace assumption are necessary and sufficient conditions for SRC to achieve high classification accuracy \cite{RigamontiBrownLepetit2011, ZhangYangFeng2011, ShiEriksson2011, ChiPorikli2013}. 
This motivates us to further investigate SRC and the underlying theoretical properties. In this paper we  propose a new implementation of SRC via screening, which is much faster than via $\ell 1$ minimization with comparable numerical performance. We further prove its theoretical consistency properties on random graphs realized from latent position models, specifically the stochastic blockmodels. 
Our results make SRC more appealing in terms of theoretical foundation, computational complexity and general applicability.

Our contributions are summarized as follows:
\begin{itemize}
    \item We are the first to propose sparse representation classification via screening to significantly reduce run time while maintaining classification efficacy and prove theoretical guarantee on its consistency for random graph models. 
    \item We extend sparse representation classification beyond $\ell 1$ minimization to achieve variable selection. We analyze the differences of SRC and SRC via screening and establish their equivalence under regularity conditions. 
    \item We demonstrate in simulation and real world networks that SRC via screening has competitive vertex classification performance but significantly faster compared with SRC via $\ell 1$ minimization. 
\end{itemize}


\section{Background and Definitions}
\label{sec:background}
Denote the training data matrix by $\mathcal{X}=[x_{1},x_{2},\ldots,x_{n}] \in \mathbb{R}^{m \times n}$, the class label vector by $\mathcal{Y}=[y_{1},y_{2},\ldots,y_{n}] \in [K]^{n}$, where $m$ is the feature dimension, $n$ is the number of observations, and $K$ is the number of classes and $[K]=[1, \ldots, K]$. A common statistical framework assumes that $(x,y),(x_1,y_1),\cdots,(x_n,y_n)$ are independent realizations from the distribution $F_{XY}$, where $(x,y) \in \mathbb{R}^{m} \times [K]$ is the testing pair and $y$ is the true but unobserved label. A classifier $g_{n}(x,D_{n})$ is a function that estimates the unknown label $y \in [K]$ based on the training pairs $D_{n}=\{(x_1,y_1),\cdots,(x_n,y_n)\}$ and the testing observation $x$. For brevity, we always denote the classifier as $g_{n}(x)$, and the classifier is correct when $g_{n}(x) = y$. Throughout the paper, 
we assume all observations are of unit norm ($\|x_{i}\|_{2}=1$), because SRC scales all observations to unit norm by default.

The sparse representation selects a subset of the training data that best represents the testing observation. Suppose $s$ is the sparsity level, we denote the subset of training data as $\widehat{\mathcal{X}}=[x_{(1)},x_{(2)},\ldots,x_{(s)}] \in \mathbb{R}^{m \times s}$. Once $\widehat{\mathcal{X}}$ is determined, $\hat{\beta}$ is the $s \times 1$ least square regression coefficients between $\widehat{\mathcal{X}}$ and $x$, and the regression residual equals $\|x-\widehat{\mathcal{X}} \hat{\beta}\|_{2}$. For each $k \in [K]$ and a given $\widehat{\mathcal{X}}$, we define
\begin{align*}
& \widehat{\mathcal{X}}_{k}=\{x_{(i)} \in \widehat{\mathcal{X}}, i=1,\ldots,s \ | \ y_{(i)}=k\} \\
& \widehat{\mathcal{X}}_{-k}=\{x_{(i)} \in \widehat{\mathcal{X}}, i=1,\ldots,s \ | \ y_{(i)} \neq k\}.
\end{align*}
Namely, $\widehat{\mathcal{X}}_{k}$
is the subset of $\widehat{\mathcal{X}}$ that contains all observations from class $k$, and $\widehat{\mathcal{X}}_{-k} = \widehat{\mathcal{X}} - \widehat{\mathcal{X}}_{k}$. We further denote $\hat{\beta}_{k}$ as the regression coefficients of $\hat{\beta}$ corresponding to $\widehat{\mathcal{X}}_{k}$, and $\hat{\beta}_{-k}$ as the regression coefficients corresponding to $\widehat{\mathcal{X}}_{-k}$, i.e.,
\begin{align*}
\widehat{\mathcal{X}}_{k} \hat{\beta}_{k}+ \widehat{\mathcal{X}}_{-k} \hat{\beta}_{-k}= \widehat{\mathcal{X}} \hat{\beta}.
\end{align*}
The original SRC makes use of the class-wise regression residual $\|x-\widehat{\mathcal{X}}_{k} \hat{\beta}_{k}\|_{2}$ in Algorithm~\ref{algSRC1}. 

\subsection*{Sparse Representation Classification by $\ell 1$}
\label{ssec:src}
SRC consists of three steps: subset selection, least square regression, and classification. Algorithm~\ref{algSRC1} describes the original algorithm: Equation~\ref{l1min} identifies the sparse representation, and solves the least square regression coefficients $\hat{\beta}$ at the same time. Then Equation~\ref{al2} assigns the class by minimizing the class-wise regression residual. Computation-wise, the $\ell 1$ minimization step takes at least $O(mns)$, while the classification step is much cheaper and takes $O(msK)$.

\begin{algorithm}
\caption{Sparse representation classification by $\ell 1$ minimization} 
\label{algSRC1}
\begin{algorithmic}[1]
\STATE \textbf{Input}: 
The training data matrix $\mathcal{X}$ with associated label vector $\mathcal{Y}$, and the testing observation $x$. 

\STATE \textbf{$\ell 1$ Minimization:} For each testing observation $x$, find $\widehat{\mathcal{X}}$ and $\hat{\beta}$ that solves the $\ell 1$ minimization problem:
\begin{equation}
\label{l1min}
\hat{\beta}=\arg\min \|\beta\|_{1}    \text{ subject to } \|x-\widehat{\mathcal{X}}\beta\|_{2} \leq \epsilon.
\end{equation} 

\STATE \textbf{Classification:} Assign the testing observation by minimizing the class-wise residual, i.e.,  
\begin{equation}
\label{al2}
g_{n}^{\ell 1}(x)=\arg\min_{k \in [K]} \|x - \widehat{\mathcal{X}}_{k} \hat{\beta}_{k}\|_{2}, 
\end{equation} 
break ties deterministically.

\STATE \textbf{Output}: The estimated class label $g_{n}^{\ell 1}(x)$. 
\end{algorithmic} 
\end{algorithm}

The $\ell 1$ minimization step is a crucial step to the success of SRC, but also the most computationally expensive step of SRC. 
There exists various iterative implementations of similar complexity, such as $\ell 1$ homotopy method \cite{OsbornePresnellTurlach2000a, OsbornePresnellTurlach2000b, EfronHastie2004}, orthogonal matching pursuit (OMP) \cite{Tropp2004, TroppGilbert2007}, augmented Lagrangian method \cite{YangZhouGanesh2013}, among many others. We use the homotopy algorithm for subsequent analysis and numerical comparison without delving into the algorithmic details. 

Note that model selection of $s$ is inherent to the $\ell 1$ minimization problem. One need to either specify a tolerance noise level $\epsilon$ or a maximum sparsity level in order for the iterative algorithm to stop.
The choice does not affect the theoretical result, but can impact the actual numerical performance and thus a separate topic for investigation \cite{Zhang2009,CaiWang2011}. 

\subsection*{Stochastic Blockmodels}
\begin{defi*}[Directed and Unweighted Stochastic Blockmodel (SBM)]
Given the class membership $\mathcal{Y}$, a directed stochastic blockmodel generates an $n \times n$ binary adjacency matrix $\mathcal{X}$ via a class connectivity matrix $V \in [0,1]^{K \times K}$ by Bernoulli distribution $\mathbf{B}(\cdot)$:
\begin{align*}
\mathcal{X}(i,j)=\mathbf{B}(V(y_i,y_j)).
\end{align*}
\end{defi*}

From the definition, the adjacency matrix produced by SBM is a high-dimensional object that is characterized by a low-dimensional class connectivity matrix. 
\subsection*{Contamination Scheme}
\begin{defi*}[Stochastic Blockmodel with Contamination]
Under the stochastic blockmodel, for each class $k$ define $V_k$ as the contamination vector of size $1 \times m$:
\begin{align*}
&V_{k}(j)=1 \mbox{ when $j$th dimension is not contaminated,}\\
&V_{k}(j)=0 \mbox{ when $j$th dimension is contaminated}.
\end{align*}
Then the contaminated random variable $X$ is
\begin{align*}
X|Y=diag(V_{Y})W_{Y}U,
\end{align*}
where $diag(V_{k})$ is a diagonal matrix of size $m \times m$ and $diag(V_{k})(j,j)=V_{k}(j)$.
\end{defi*}
In Section \ref{sec:numer}, we use the above contamination scheme to contaminate both the simulated and the real world networks and show SRC via screening achieves the same level of robustness as the original SRC algorithm.

\section{Main Results}
\label{sec:main}

In this section, we present our proposed SRC algorithm via screening, which differs from the original SRC algorithm mostly in the subset selection step and slightly in the classification step. We investigate the theoretical properties of the classification step by proving classification consistency for stochastic blockmodels.


\subsection{SRC via Screening}

Our proposed algorithm is shown in Algorithm~\ref{algSRC2}, which replaces the $\ell 1$ minimization step by screening, and minimizes the class-wise residual in angle in the classification step. Algorithm~\ref{algSRC2} obtains better computation complexity because the screening procedure becomes non-iterative and simply chooses $s$ observations out of $\mathcal{X}$ that are most correlated with the testing observation $x$ as $\widehat{\mathcal{X}}$. This step merely requires $O(mn+nlog(n))$ instead of $O(mns)$ for $\ell 1$. 

Indeed, the screening procedure has recently gained popularity as a fast alternative of regularized regression for high-dimensional data inference. The speed advantage makes it a suitable candidate for efficient data extraction, and is shown to be equivalent to $\ell 1$ and $\ell 0$ minimization under various regularity conditions \cite{FanLv2008,FanSamworthWu2009,WassermanRoeder2009,FanFengSong2011,GenoveseEtAl2012,KolarLiu2012}. In particular, it is argued that the maximal sparsity level $s=\max\{n/log(n),m\}$ works well for screening \cite{FanLv2008}. Thus we also set the same sparsity level in our experiments in Section \ref{sec:numer}.

\begin{algorithm}
\caption{Sparse representation classification via screening for vertex classification in stochastic blockmodels} 

\begin{algorithmic}
\STATE \textbf{Input}: 
The adjacency $\mathcal{X}$, the known label vector $\mathcal{Y}$, and the testing vertex $x$. 

\STATE \textbf{Screening:} Calculate $\Omega=\{x_{1}^{T} x, x_{2}^{T} x, \cdots, x_{n}^{T} x\}$ ($^{T}$ is the transpose), and sort the elements by decreasing order. Take $\widehat{\mathcal{X}}=\{x_{(1)}, x_{(2)}, \ldots, x_{(s)}\}$ with $s=\min\{n/log(n),m\}$, where $x_{(i)}^{T}x$ is the $i$th largest element in $\Omega$.

\STATE \textbf{Regression:} Solve the ordinary least square problem between $\widehat{\mathcal{X}}$ and $x$. Namely, compute $\beta=\widehat{\mathcal{X}}^{-1} x$ where $\widehat{\mathcal{X}}^{-1}$ is the Moore-Penrose inverse.

\STATE \textbf{Classification:} Assign the testing vertex by 
\begin{equation}
\label{al2}
g_{n}^{scr}(x)=\arg\min_{k \in [K]} \theta(x, \widehat{\mathcal{X}}_{k} \hat{\beta}_{k}), 
\end{equation}
where $\theta$ denotes the principal angle. Break ties deterministically.

\STATE \textbf{Output}: The estimated class label $g_{n}^{scr}(x)$. 
\end{algorithmic} 
\label{algSRC2}
\end{algorithm}

\subsection{Consistency Under Stochastic Block-Model}
Here we prove SRC consistency for the stochastic blockmodel~\cite{HollandEtAl1983,SussmanEtAl2012,Lei2015}, which is a popular network model commonly used for classification and clustering. Although the results are applicable to undirected, weighted, and other similar graph models, for ease of presentation we concentrate on the directed and unweighted SBM.

\begin{thm}
\label{thm:sbm}
Denote $\rho \in [0,1]^{K}$ as the prior probability of each class, SRC by screening is consistent for vertex classification under SBM when 
\begin{align}
\label{eq:sbm}
\frac{\rho \circ V(Y,:)V(Y,:)^{T}}{\|\rho \circ V(Y,:)\|_{1}^{1/2}} > \frac{\rho \circ V(Y,:)V(Y',:)^{T}}{\|\rho \circ V(Y',:)\|_{1}^{1/2}}
\end{align}
holds for $Y' \neq Y$, where $\circ$ denotes the entry-wise product.
\end{thm}
The condition also guarantees data of the same class to be more similar in angle than data of different classes, thus inherently the same as the principal angle condition for latent subspace mixture model.

\section{Numerical Experiments}
\label{sec:numer}
In this section we apply the SRC screening algorithm to simulated and real world network graphs. The evaluation criterion is the leave-one-out error for vertex classification: within each vertex set, one vertex is held out for testing and the remaining are used for training, do the classification, and repeat until each observation in the given data is hold-out once. SRC is shown to be a robust vertex classifier in \cite{ChenShenVogelsteinPriebe2016} with superior performance to other classifiers for both simulated and real networks. We use the SRC algorithm's performance for vertex classification as a baseline for comparison. 

Our simulation and real data experiments show that SRC is consistent under stochastic blockmodel and robust against contamination on graphs. Overall, we observe that SRC by screening performs very similar to SRC without screening, but with much faster run time for training.

\subsection{Stochastic Blockmodel Simulation}
\begin{figure}
\centering
\subfloat[]{
\includegraphics[width=0.25\textwidth]{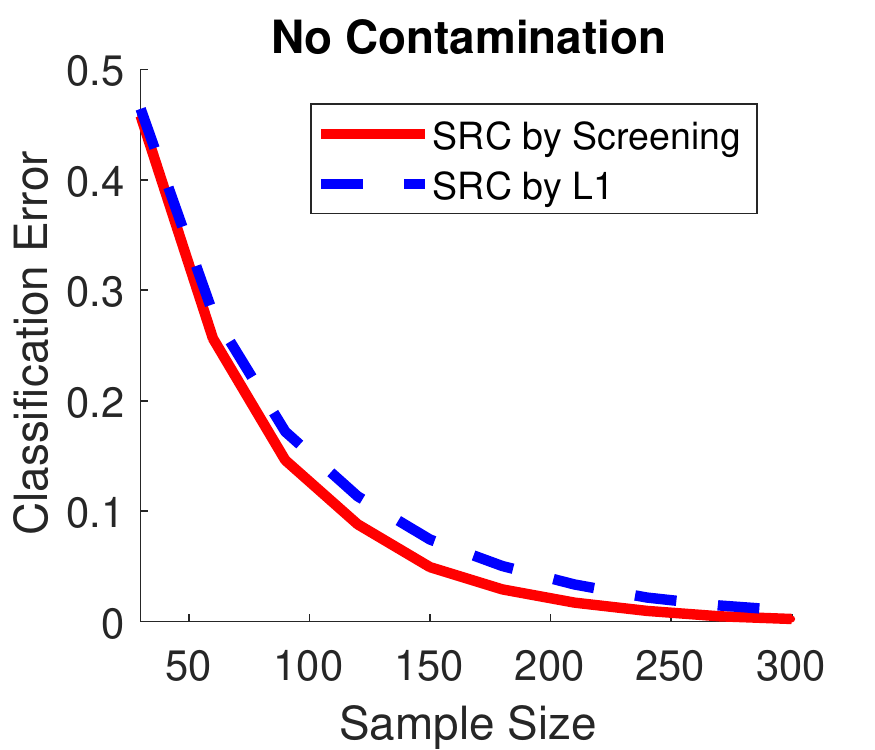}
}
\subfloat[]{
\includegraphics[width=0.25\textwidth]{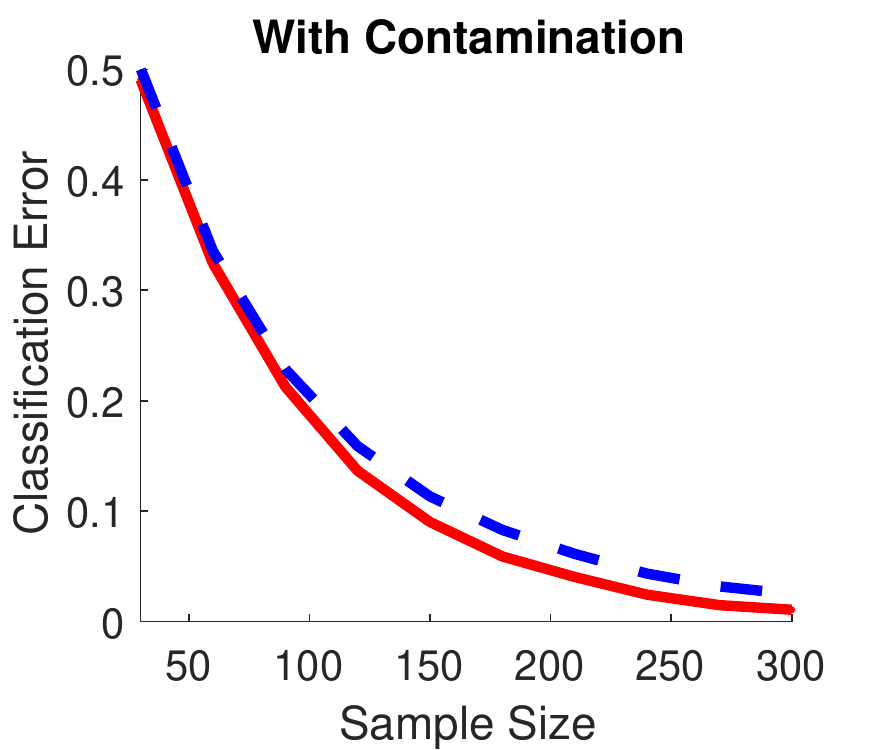}
}
\caption{SRC screening achieves the same classification error as SRC for vertex classification on simulated graphs. }
\label{figSRC1}
\end{figure}

We simulate graphs from the stochastic blockmodel with number of blocks $K=3$, membership prior $(\rho_{1}=\rho_{3}=0.3, \rho_{2}=0.4)$ and the class connectivity matrix $V$ where
\begin{eqnarray*}
    V=
    \begin{bmatrix}
    0.3       & 0.1 & 0.1 \\
    0.1       & 0.3 & 0.1  \\
    0.1       & 0.1 & 0.3
\end{bmatrix}.
\end{eqnarray*}
The class connectivity matrix satisfies the condition in Theorem~\ref{thm:sbm}. From the SBM, we simulate graphs with varied sizes for $n=30,60,\ldots,300$, compute the leave-one-out error, then repeat for $100$ Monte-Carlo replicates and plot the average errors in Figure~\ref{figSRC1}. 

\begin{table}
\centering
\caption{Leave-one-out error and run time (s) comparison}
\label{t:real1}%
\begin{tabular}{|c|c|c|c|}
\hline
Algorithm  & Metrics& Wikipedia & \textit{C.elegans}  \\
\hline
SRC by Screening & Error &$32.27 \%$ & $42.29 \%$ \\
\hline
SRC by $\ell 1$ & Error & $29.31 \%$ & $48.62 \%$ \\
\hline
SRC by Screening & Runtime & $32.3$ & $0.3$ \\
\hline
SRC by $\ell 1$ & Runtime  & $573.7$ & $9.1$ \\
\hline
\end{tabular}
\end{table}




\begin{figure}
\subfloat[]{
\includegraphics[width=0.25\textwidth]{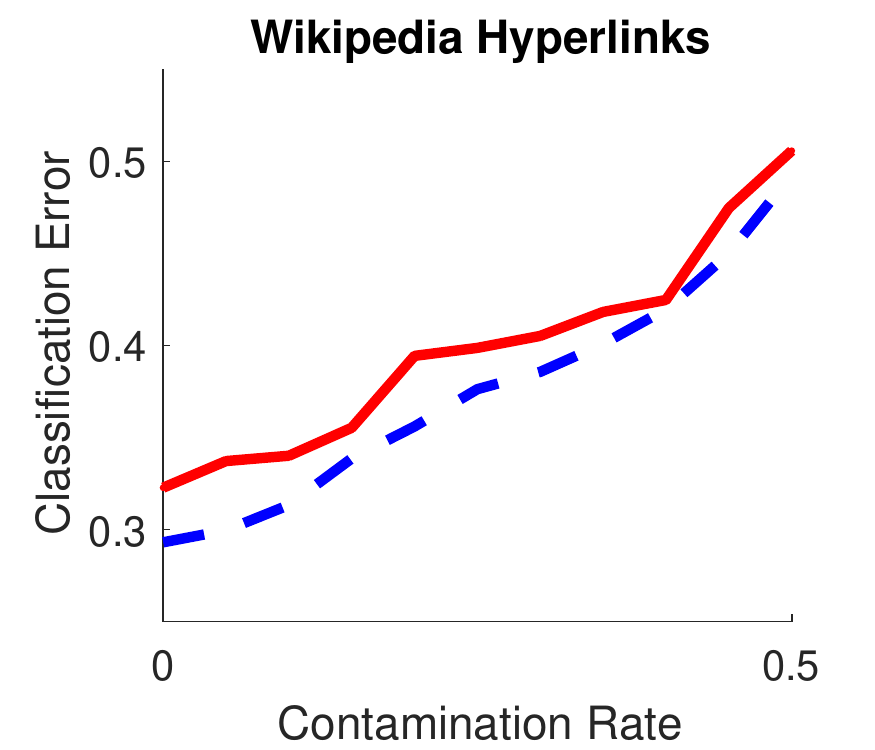}
}
\subfloat[]{
\includegraphics[width=0.25\textwidth]{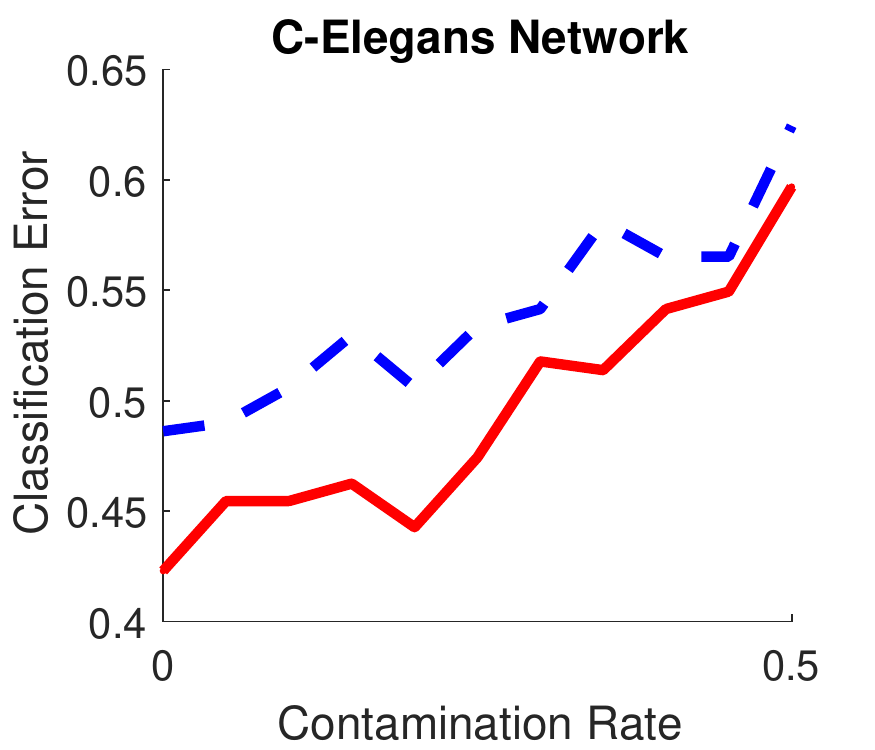}
}
\caption{SRC on contaminated real networks. SRC via screening and SRC have similar robustness against contamination. On the \textit{C.elegans} neuro-connectome, SRC via screening maintains more robustness than SRC in terms of lower error, as the contamination rate increases. }
\label{figSRC2}
\end{figure}

\subsection{Network Connectivity}

Next we apply SRC to vertex classification on real-world networks. The first graph is collected from Wikipedia article hyperlinks \cite{PriebeMarchette2012}. A total of $1382$ English documents based on the 2-neighborhood of the English article ``algebraic geometry'' are collected, and the adjacency matrix is formed via the documents' hyperlinks. This is a directed, unweighted, and sparse graph without self-loop, where the graph density is $1.98\%$. There are five classes based on article contents: $119$ articles in category class, $372$ articles about people, $270$ articles about locations, $191$ articles on date, and $430$ articles are real math.  

The second graph data is the electric neural connectome of Caenorhabditis elegans (\textit{C.elegans}) \cite{hall1991posterior,varshney2011structural,chen2015celegans}. The hermaphrodite \textit{C.elegans} somatic nervous system has over two hundred neurons, classified into $3$ classes: motor neurons, interneurons, and sensory neurons. The adjacency matrix is also undirected, unweighted, and sparse with density $1.32\%$. 

The leave-one-out errors and run times are reported in Table~\ref{t:real1}. On the Wikipedia graph, SRC with screening has slightly higher error rate at 32.27\% than SRC without screening at 29.31\%. On the \textit{C.elegans} neuro-connectome, the SRC with screening is much lower at 42.29\% than SRC without screening at 48.62\%.
We further conducted contamination analysis on these two real networks. The contaminated classification performance are shown in the bottom panels of Figure~\ref{figSRC2}, where the solid lines are performance by SRC via screen and the dotted lines are performance by the original SRC. SRC via screening demonstrated more robustness in terms of lower classification error, compared with SRC without screening on the  \textit{C.elegans} neuro-connectome.
\section{Conclusion}
In this paper, we propose sparse representation classification via screening for random graphs distributed as stochastic blockmodels. We prove the theoretical consistency and show in simulation and real world experiments the effectiveness and robustness of our proposed algorithm. We continue our research of extending SRC via screening to latent subspace mixture models and quantifying robustness against contamination. 


\bibliography{references}
\bibliographystyle{icml2019}

\section*{Appendix: Sketch of proof}
\label{sec:pf}

\subsection*{Theorem~\ref{thm:sbm}}
\begin{sproof}

Denote $x$ as the testing adjacency vector of size $1 \times n$, and $x_{1}$ as a training adjacency vector of size $1 \times n$. Then 
\begin{align*}
\cos \theta (x,x_{1})= &\frac{\sum_{j=1}^{n} \mathcal{B}(V(y,Y_{j})V(y_{1},Y_{j}))}{\sqrt{\sum_{j=1}^{n} \mathcal{B}(V(y,Y_{j})) \sum_{j=1}^{n} \mathcal{B}(V(y_{1},Y_{j}))}}\\
\stackrel{n \rightarrow \infty}{\rightarrow} &\frac{ E(V(y,Y)V(y_{1},Y))}{E(V(y,Y)) E(V(y_{1},Y))}\\
= &\frac{ \sum_{k=1}^{K} \rho_{k}V(y,k)V(y_{1},k)}{\sqrt{\sum_{k=1}^{K} \rho_{k}V(y,k) \sum_{k=1}^{K} \rho_{k}V(y_{1},k)}} \\
= & \frac{\rho \circ V(y,:)V(y_{1},:)^{T}}{\sqrt{\|\rho \circ V(y,:)\|_{1}\|\rho \circ V(y_{1},:)\|_{1}}}
\end{align*}
It follows that when Equation~\ref{eq:sbm} holds, $\cos \theta (X,X') > \cos \theta (X,X_{1})$ always holds asymptotically for $Y=Y' \neq Y_{1}$. 
\end{sproof}




\end{document}